\documentclass[10pt,journal,compsoc]{IEEEtran}
\usepackage{microtype}
\usepackage{graphicx}
\usepackage{subfigure}
\usepackage{booktabs} 
\usepackage{threeparttable}
\usepackage{amsmath}
\usepackage{amssymb}
\usepackage{mathtools}
\usepackage{amsthm}
\usepackage{multirow}
\usepackage{makecell}
\usepackage{bm}

\ifCLASSOPTIONcompsoc
  \usepackage[nocompress]{cite}
\else
  \usepackage{cite}
\fi
\ifCLASSINFOpdf
\else
\fi
\begin{document}
%
\title{Torsion Graph Neural Networks}
%
%
%
%

\author{Cong Shen,~Xiang Liu,~Jiawei Luo and Kelin Xia
\IEEEcompsocitemizethanks{\IEEEcompsocthanksitem Cong Shen and Jiawei Luo are with College of Computer Science and Electronic Engineering, Hunan University, Changsha, China, 410000. E-mail: \{cshen,luojiawei\}@hnu.edu.cn. Cong Shen is also with School of Physical and Mathematical Sciences, Nanyang Technological University, Singapore, 637371. 
\IEEEcompsocthanksitem Xiang Liu is with Chern Institute of Mathematics and LPMC, Nankai University, Tianjin, China, 300071. E-mail: liuxiangmath@163.com. Xiang Liu is also with School of Physical and Mathematical Sciences, Nanyang Technological University, Singapore, 637371.
\IEEEcompsocthanksitem Kelin Xia is with School of Physical and Mathematical Sciences, Nanyang Technological University, Singapore, 637371. E-mail: xiakelin@ntu.edu.sg}
\thanks{Corresponding author: Jiawei Luo and Kelin Xia}
\thanks{Manuscript received April 19, 2005; revised August 26, 2015.}}

%
%

\markboth{Journal of \LaTeX\ Class Files,~Vol.~14, No.~8, August~2015}%
{Shell \MakeLowercase{\textit{et al.}}: Bare Demo of IEEEtran.cls for Computer Society Journals}
%



\IEEEtitleabstractindextext{%
\begin{abstract}
Geometric deep learning (GDL) models have demonstrated a great potential for the analysis of non-Euclidian data. They are developed to incorporate the geometric and topological information of non-Euclidian data into the end-to-end deep learning architectures. Motivated by the recent success of discrete Ricci curvature in graph neural network (GNNs), we propose TorGNN, an analytic Torsion enhanced Graph Neural Network model. The essential idea is to characterize graph local structures with an analytic torsion based weight formula. Mathematically, analytic torsion is a topological invariant that can distinguish spaces which are homotopy equivalent but not homeomorphic. In our TorGNN, for each edge, a corresponding local simplicial complex is identified, then the analytic torsion (for this local simplicial complex) is calculated, and further used as a weight (for this edge) in message-passing process. Our TorGNN model is validated on link prediction tasks from sixteen different types of networks and node classification tasks from three types of networks. It has been found that our TorGNN can achieve superior performance on both tasks, and outperform various state-of-the-art models. This demonstrates that analytic torsion is a highly efficient topological invariant in the characterization of graph structures and can significantly boost the performance of GNNs.
\end{abstract}

}

\maketitle

\IEEEdisplaynontitleabstractindextext

%
\IEEEpeerreviewmaketitle

\IEEEraisesectionheading{\section{Introduction}\label{sec:introduction}}
\IEEEPARstart{W}{ith} the accumulation of non-Euclidean data, the development of deep learning models, which have revolutionized sequence and image data analysis \cite{lecun2015deep}, has given rise to geometric deep learning (GDL). Among all the non-Euclidean data are graphs and networks, which are arguable the most powerful topological representation of real-world data and systems. As a special type of GDL, graph neural networks (GNNs) have demonstrated remarkable learning capability and have become prevalent models for various graph tasks, such as node classification, link prediction, and graph classification \cite{welling2016semi,hamilton2017inductive,vashishth2019composition,velivckovic2017graph}. GNNs have achieved great success in applications, such as molecule property prediction \cite{fout2017protein}, recommender systems \cite{fan2019graph}, natural language processing \cite{young2018recent}, critical data classification \cite{abu2020n}, computer vision \cite{shen2020auto}, particle physics \cite{ju2020graph}, and resource allocation in computer networks \cite{rusek2018message}. Recent years have seen a rapid increase of research in the field of GNNs. Great efforts have been devoted to  algorithm efficiency improvement (especially for large graphs), special architecture design, and various applications \cite{abadal2021computing}.

In general, GNN models can be divided into several types, including recurrent-based GNNs, convolution-based GNNs, graph autoencoders, graph reinforcement learning, and graph adversarial networks \cite{abadal2021computing,zhou2020graph}. The recurrent-based GNNs refer to the initial GNN models, which employ recurrent units as its combination functions. Typical examples are CommNet \cite{sukhbaatar2016learning} and GG-NN \cite{li2015gated}. The convolution-based GNNs expand the idea of convolution in the graph space to graph spectral space, based on spectral graph theory. These models are computationally much more affordable, flexible, and scalable. Examples of these models include GCN \cite{welling2016semi}, CurvGN \cite{ye2019curvature}, FastGCN \cite{chen2018fastgcn}, Cluster-GCN \cite{chiang2019cluster}, LGCL \cite{gao2018large}, ST-GCN \cite{yan2018spatial}, AGCN \cite{li2018adaptive} and SGCs \cite{wu2019simplifying}. Graph Autoencoder (GAE) is a different type of GNNs, which converts the graph structure into a latent representation (i.e., encoding), that can be later expanded to a graph structure as close as possible to the original one (i.e., decoding). NetRA \cite{yu2018learning} is a classical GAE with a good performance. Finally, graph reinforcement learning and graph adversarial networks are newly-proposed GNNs that combine reinforcement learning and generative confrontation networks with graph neural architecture. Typical models include MolGAN \cite{de2018molgan} and MINERVA\cite{das2017go}. In terms of application, recurrent-based GNNs are mainly used in sequence data; Convolution-based GNNs are mainly used in various graphs, networks or knowledge graphs; Graph autoencoders are mainly used in the field of unsupervised models, which are suitable for feature selection or dimensionality reduction; Graph reinforcement learning and graph adversarial networks are applied to generative models, such as molecular generation.

\begin{figure}[ht]
\begin{center}
\includegraphics[width=3.2in]{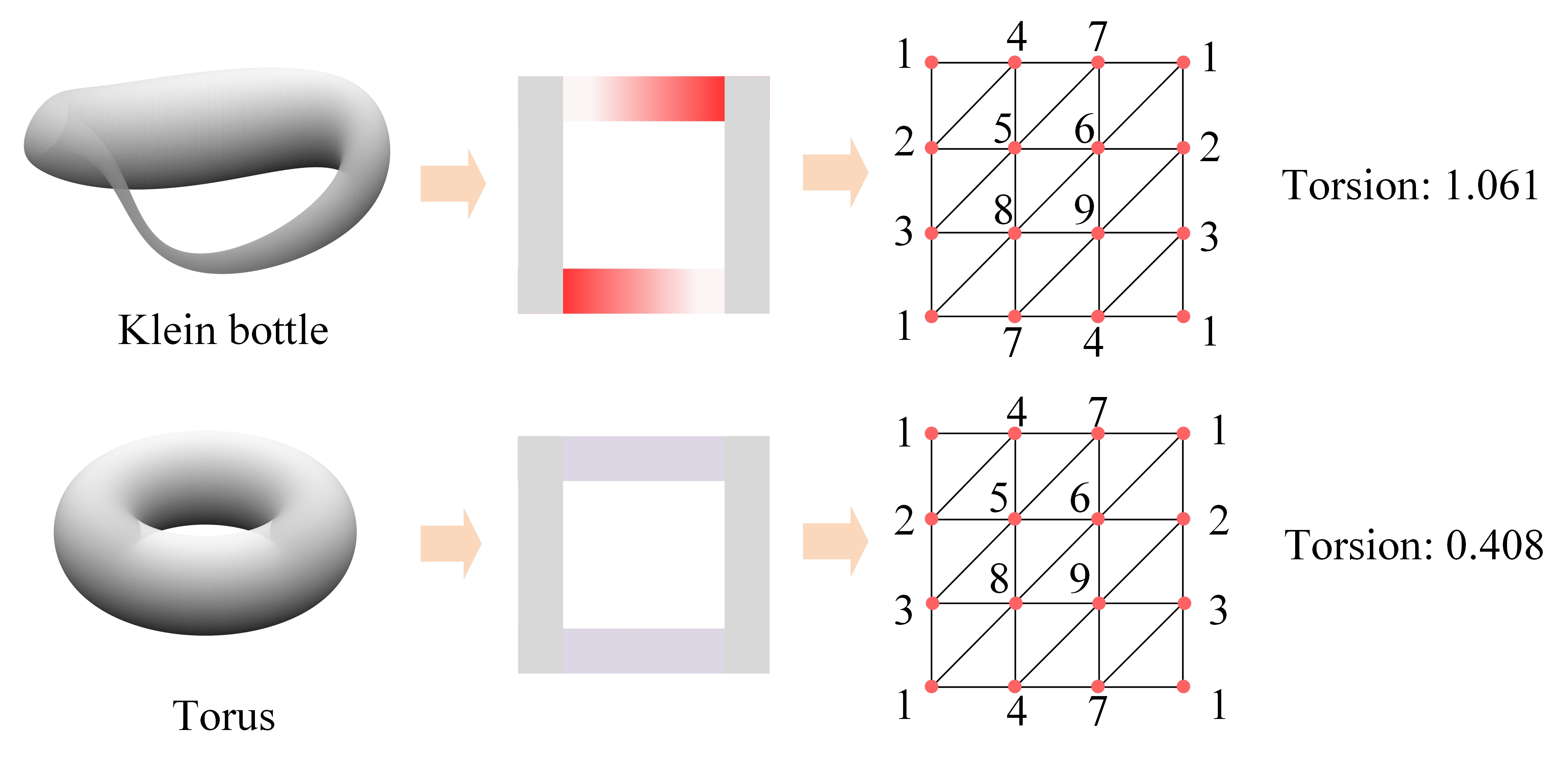}
\caption{\textbf{The illustration of characterization capability of analytic torsion for Klein bottle and torus surface.} Both Klein bottle and torus surface can be obtained by ``gluing" together their edges. A twist exists in Klein bottle when its upper-side is ``glued" with low-side as indicated by red colors, i.e., regions in upper edge and low edge are glued together when they are of the same color. More specifically, the two manifolds can be discretized into two simplicial complexes with vertices of same numbers gluing together. Using 1D Hodge Laplacians, analytic torsion for Klein bottle and torus are 1.061 and 0.408, respectively.}\label{fig:Klein bottle and Torus}
\end{center}
\end{figure}

Even with the great development and progress in GNNs, the incorporation of geometric and topological information into graph neural architecture is still a key issue for all GNN and GDL models. Recently, discrete Ricci curvatures have been used in the characterization of ``over-squashing" phenomenon \cite{topping2021understanding}, which happens at the bottleneck region of a network where the messages in GNN models propagated from distant nodes distort significantly. Curvature-based GNNs have been developed by the incorporation of Ollivier Ricci curvature, a discrete Ricci curvature model, into GNN models and have achieved great success in various synthetic and real-world graphs, from social networks, coauthor networks, citation networks, and Amazon co-purchase graphs \cite{ye2019curvature,li2022curvature}. The model can significantly outperform state-of-the-art models when the underlying graphs are of large-sized and dense \cite{ye2019curvature,li2022curvature}.

Analytic torsion, which is an alternating product of determinants of Hodge Laplacians, is a topological invariant which can distinguish spaces that are homotopy equivalent but not homeomorphic \cite{turaev2001introduction,grigor2020torsion}. Analytic torsion is introduced as the analytic version of Reidemeister torsion (or R-torsion for short), which is an algebraic topology invariant that takes values in the multiplicative group of the units of a commutative ring \cite{ray1971r}. R-torsion was developed for the classification of three dimensional (3D) lens spaces in 1935 by Reidemeister \cite{reidemeister1935homotopieringe}. It is found that a complete classification of 3D lens spaces can be achieved in terms of R-torsion and fundamental group \cite{turaev2001introduction}. To describe the R-torsion in analytic terms, Ray and Singer defined the analytic torsion for any compact oriented manifold and orthogonal representation of the fundamental group \cite{ray1971r}. Their definition uses the spectrum of the Hodge Laplacian on twisted forms. When the orthogonal representation is acyclic and orthogonal, Cheeger and M\"{u}ller proves that the R-torsion is equivalent to analytic torsion \cite{muller1978analytic, cheeger1977analytic}. With the unique characterization capability, analytic torsion provides a powerful representation of the topological and geometric information within the data. As demonstrated in Figure \ref{fig:Klein bottle and Torus}, both Klein bottle and torus surface can be obtained by ``gluing" together the upper edge with lower one and left edge with right one. However, a twist exists in Klein bottle when its upper-side is glued with low-side as indicated by red colors, i.e., regions in upper edge and low edge are glued together when they are of the same color. This can be seen more clearly if the manifolds are discretized into two simplicial complexes with vertices of same numbers gluing together. Interestingly, the topological difference between Klein bottle and torus surface can be well characterized by their analytic torsion. The unique characterization capability of analytic torsion makes it an efficient topological invariant for GNNs, and more general GDLs.

Here we propose a new graph neural network model called analytic Torsion enhanced  Graph Neural Network (TorGNN). An analytic torsion based message passing process is developed in our TorGNN. Mathematically, a local simplicial complex is constructed for each edge, and its analytic torsion is used as the weight of the special edge in node feature aggregation. Our TorGNN model shows the best performance and outperforms all state-of-the-art methods, on link prediction tasks from 16 different datasets and node classification tasks from 3 different datasets. This demonstrate that our TorGNN can better capture the complexity of the local structure of graph data.

\section{Related work}
\subsection{Geometry-aware graph neural networks}
Geometry-aware GNNs have been proposed to incorporate algebraic and geometric information of graph data, though the refined message passing and aggregation mechanisms. Among them is TFN, which is a SE(3) equivariant GNN model based on the group set of 3D translation and rotation transformations \cite{thomas2018tensor}. LieConv is based on Lie group set of differential transformations beyond 3D translations and rotations \cite{finzi2020generalizing}. EGNN makes use of all $n$-dimension Euclidean transformations including translations, rotations and reflections \cite{satorras2021n}.

In addition, other GNN models take into account the curvature information, such as CurvGN  \cite{ye2019curvature}, SELFMGNN \cite{sun2022self} and CurvGAN \cite{li2022curvature}. CurvGN is the first graph convolutional network built on advanced graph curvature information. SELFMGNN is the first attempt to study the self-supervised graph representation learning in the mixed-curvature spaces. CurvGAN is the first GAN-based graph representation method in the Riemannian geometric manifold. Curvature has also been combined with hyperbolic graph neural networks, such as, HGCN \cite{chami2019hyperbolic}, HAT \cite{zhang2021hyperbolic}, HGNN \cite{liu2019hyperbolic}, ACE-HGNN \cite{fu2021ace} and HRGCN$+$ \cite{wu2021hyperbolic}.

\subsection{Analytic Torsion}
The Reidemeister torsion, or R-torsion, was proposed by Kurt Reidemeister in 1935 \cite{reidemeister1935homotopieringe}. It is the first topological invariant that could distinguish spaces with the same homotopy type. With fundamental group, it has been used in the complete classification of 3D lens spaces. In 1970s, Ray and Singer introduced the analytic torsion as the analytic version of Reidemeister torsion for any compact oriented manifold and orthogonal representation of the fundamental group \cite{ray1971r}.  When the orthogonal representation is acyclic and orthogonal, Cheeger and M\"{u}ller proves that the R-torsion is equivalent to analytic torsion \cite{muller1978analytic, cheeger1977analytic}. Computationally, analytic torsion can be calculated from the determinants of Hodge Laplacians \cite{grigor2020torsion}. Note that for Hodge Laplacians with zero eigenvalues, the determinant is replaced by the multiplication of all non-zero eigenvalues. With the unique characterization capability, analytic torsion provides a powerful representation of the topological and geometric information within the data. As far as we know, this is the first attempt to combine analytic torsion and GNNs.

\section{TorGNN}
\subsection{Notations and Problem Formulation}
\textbf{Notations}
Let $G = (V,E)$ represents a graph with nodes $v_i \in V$ ($x$ and $y$ are also used to denote nodes) and edges $\left(v_i, v_j\right) \in E$. Node features are denoted as $H = \left\{ {h_{1},\cdots,h_{N}} \right\} \in \mathbb{R}^{N \times m}$. A total number of $N$ nodes in vertex set $V$ and each node is encoded with a predefined $m$-dimension attribute vector (e.g., a generated graph embedding or a one-hot coding). We use $\mathcal{N}(x)$ to denote the neighbors of node $x$, $d(x)$ represent node degree of nodes $x$.

We use $K$ to represent a simplicial complex, $K_{x,y}$ the local simplicial complex from nodes $x$ and $y$, and $T(K_{x,y})$ the analytic torsion for local simplicial complex $K_{x,y}$. The $p$-th Hodge Laplacian is denoted as $L_p$ and its determinant is denoted as $|L_p|$. If the matrix has zero eigenvalues, the determinant $|L_p|$ equals to the multiplication of all non-zero eigenvalues. The $p$-th boundary matrix is denoted as $B_p$. Zeta function can be defined based on the eigenvalues of $L_{p}$ and is denoted as $\zeta_{p}(s)$.

\textbf{Problem formulation}
We consider two types of tasks, one is link prediction and the other is node classification. The link prediction task is to learn a mapping function $\left. \Phi: E \rightarrow\lbrack 0,1\rbrack \right.$ from edges to scores, such that we can obtain the probability of two arbitrary nodes interacting with each other. Similarity, node classification is to learn a mapping function $\left. \Psi: V\rightarrow \mathbb{R}^N_{c} \right.$ from nodes to vectors and $N_{c}$ is dimension of node's labels.


\subsection{Analytic torsion}

\begin{figure*}[h]
\begin{center}
\includegraphics[width=5.5in]{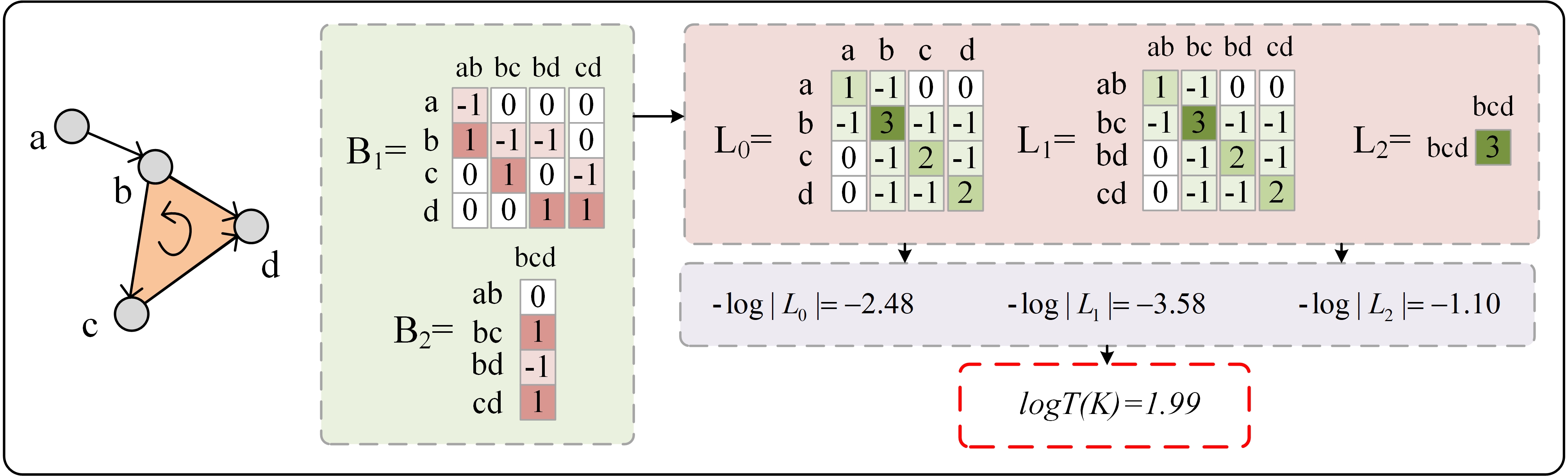}
\caption{\textbf{An example of calculating the logarithm of analytic torsion for a two dimension simplicial complex.} Based on the simplicial complex $K$, we can construct boundary matrices, namely $B_1$ and $B_2$. The Hodge Laplacian matrices, including $L_0$, $L_1$ and $L_2$, can be calculated according to Eq. (\ref{eq:L}). Then, the determinant of Hodge Laplacian matrices $|L_{p}|$ can be evaluated. Finally, the logarithm of the analytic torsion $\log T(K)$ can be calculated.  }\label{fig:example of analytic torsion}
\end{center}
\end{figure*}

\subsubsection{Simplicial complex}
As a generalization of graph, a simplicial complex $K$ is composed of simplices. A simplex with $p+1$ vertices from a vertex set $V$ is called a $p$-simplex, denoted by $\sigma^{p} = \left\{ {v_{0},v_{1},\cdots,v_{p}} \right\}$. For a $p$-simplex $\sigma^{p}$, any nonempty subset is called its face. A face of $\sigma^{p}$ with $p-1$ dimension is called a boundary of $\sigma^{p}$. Geometrically, a $p$-simplex can be seen as the convex hull formed by $p+1$ affinely independent points. In this way, $0$-simplex is a vertex, $1$-simplex is an edge, $2$-simplex is a triangle, and $3$-simplex is a tetrahedron.

\subsubsection{Combinatorial Laplacian of simplicial complex}
Given an oriented simplicial complex $K$, that is, a simplicial complex with an order on the vertex set, we use ${Z/2}$ coefficient and denote a $p$-simplex as $\sigma^{p} = \left\lbrack v_{0},v_{1},\cdots,v_{p} \right\rbrack$. An $p$-chain is defined as a finite sum of $p$-simplices in $K$, denoted as $c = {\sum\limits_{i}\sigma_{i}^{p}}$. Let $C_p$ be the set of $p$-chains of $K$, which is spanned by $p$-simplices of $K$ over ${Z/2}$, called the $p$-chain group of $K$. The boundary operator $\left. \partial_{p}:C_{p}\rightarrow C_{p - 1} \right.$ for a $p$-simplex $\sigma^{p}$ is defined as follows,
\begin{align}  \nonumber
\partial_{p}\sigma^{p} = {\sum\limits_{i = 0}^{p}\left( {- 1} \right)^{i}}\left\lbrack {v_{0},\cdots,{\hat{v}}_{i},\cdots,v_{p}} \right\rbrack
\end{align}
where ${\hat{v}}_{i}$ means that $v_i$ has been removed. The boundary operator satisfies $\partial_{p - 1}\partial_{p} = 0$. These chain groups together with boundary operators form a chain complex,
\begin{align}  \nonumber
0\overset{\partial_{0}}{\leftarrow}C_{0}\overset{\partial_{1}}{\leftarrow}\cdots\overset{\partial_{p}}{\leftarrow}C_{p}\overset{\partial_{p + 1}}{\leftarrow}\cdots
\end{align}
Considering the canonical inner product on chain groups, that is, let all the simplices orthogonal. Denote the adjoint of $\partial_{p}$ by $\delta^{p}$, then the $p$-th Hodge (combinatorial) Laplacian operator of $K$ is defined as,
\begin{align}  \nonumber
\Delta_{p} = \delta^{p}\partial_{p} + \partial_{p + 1}\delta^{p + 1}.
\end{align}

Let $B_p$ be the matrix form of $\partial_{p}$, then, the $p$-th Hodge Laplacian matrix of $K$ is,
\begin{align}\label{eq:L}
L_{p} = B_{p}^{T}B_{p} + B_{p + 1}B_{p + 1}^{T}.
\end{align}

For $1$-dimension simplicial complex $K$, i.e., a graph, we have $L_{0} = B_{1}B_{1}^{T}$, which is the graph Laplacian. Further, for an $n$-dimension simplicial complex $K$, its highest order Hodge Laplacian is $L_{n} = B_{n}^{T}B_{n}$.

The Hodge Laplacian matrices in Eq.(\ref{eq:L}) can be explicitly described in terms of simplex relations. More specifically, $L_0$ can be expressed as,
\begin{align}  \nonumber
L_{0}\left( {i,j} \right) = \left\{ \begin{matrix}
{d\left( \sigma_{i}^{0} \right),~~{\rm if}~i = j~~~~~~~~~~~~~~~~~~} \\
{- 1,~~~~~{\rm if}~i \neq j~{\rm and} {~\sigma}_{i}^{0} \cap \sigma_{j}^{0}} \\
{0,~~~~~~{\rm if}~i \neq j~{\rm and}{~\sigma}_{i}^{0}\not{\cap}\sigma_{j}^{0}} \\
\end{matrix} \right.
\end{align}
where $d\left( \sigma_{i}^{0} \right)$ is the degree of vertex $\sigma_{i}^{0}$. Furthermore, when $p > 0$, $L_p$ can be expressed as
\begin{align}  \nonumber
L_{p}\left( {i,j} \right) = \left\{ \begin{matrix}
\begin{matrix}
{d\left( \sigma_{i}^{p} \right) + p + 1,~~{\rm if}~i = j~~~~~~~~~~~~~~~~~~~~~~~~~~~~~~~~~~} \\
{1,{\rm if}~i \neq j,{~\sigma}_{i}^{p}\not{\cap}\sigma_{j}^{p},{~\sigma}_{i}^{p} \cup \sigma_{j}^{p}~{\rm and}~\sigma_{i}^{p} \sim \sigma_{j}^{p}~} \\
\end{matrix} \\
\begin{matrix}
{-1,{\rm if}~i \neq j,{~\sigma}_{i}^{p}\not{\cap}\sigma_{j}^{p},{~\sigma}_{i}^{p} \cup \sigma_{j}^{p}~{\rm and}~\sigma_{i}^{p} \nsim \sigma_{j}^{p}} \\
{0,~~{\rm if}~i \neq j,{~\sigma}_{i}^{p} \cap \sigma_{j}^{p}~{\rm or}{~\sigma}_{i}^{p}\not{\cup}\sigma_{j}^{p}~~~~~~~~~~~~~~~~~} \\
\end{matrix} \\
\end{matrix} \right.
\end{align}
here $d\left( \sigma_{i}^{p} \right)$ is (upper) degree of $p$-simplex $\sigma_{i}^{p}$. It is the number of $(p+1)$-simplexes, of which $
\sigma_{i}^{p}$ is a face. Notation ${~\sigma}_{i}^{p} \cap \sigma_{j}^{p}$ represents that two simplexes are upper adjacent, i.e. they are faces of a common $(p+1)$-simplex, and $\sigma_{i}^{p}\not{\cap}\sigma_{j}^{p}$ represents the opposite. Notation $\sigma_{i}^{p} \cup \sigma_{j}^{p}$ represents that two simplexes are lower adjacent, i.e. they share a common $(p-1)$-simplex as their face, and ${\sigma}_{i}^{p} \not{\cup} \sigma_{j}^{p}$ represents the opposite. Notation $
\sigma_{i}^{p} \sim \sigma_{j}^{p}$ represents that two simplexes have the same orientation, i.e. oriented similarly, and $\sigma_{i}^{p} \nsim \sigma_{j}^{p}$ represents the opposite. The eigenvalues of combinatorial Laplacian matrices are independent of the choice of the orientation.

The Laplacian matrix has various important properties. First, it is always positive semi-definite, thus all its eigenvalues are non-negative. Second, the multiplicity of zero eigenvalues, i.e., the total number of zero eigenvalues, of $L_p$ is equal to the $p$-th Betti number $\beta_{p}$. In particular, If $K$ is a graph, the number (multiplicity) of zero eigenvalues is equal to the topological invariant $\beta_{0}$, which counts the number of connected components in the graph. Third, the second smallest eigenvalue, i.e., the first non-zero eigenvalue, is called Fiedler value or algebraic connectivity, which describes the general connectivity. The corresponding eigenvector can be used in classification and clustering.

\subsubsection{Analytic torsion for simplicial complex}
For an $n$-dimension simplicial complex $K$, its $p$-th Laplacian matrix is $L_{p} = B_{p}^{T}B_{p} + B_{p + 1}B_{p + 1}^{T}$ where $B_p$ is the boundary matrix. Since $L_{p}$ is a positive semi-definite, all of its eigenvalues, denoted as $\left\{ \lambda_{i} \right\}$, are non-negative. A special Zeta function can be defined based on the eigenvalues of $L_{p}$ as follows,
$$\zeta_{p}(s) = {\sum_{\lambda_{i} > 0}\frac{1}{\lambda_{i}^{s}}}.$$
Note that the Zeta function is composed of all the positive eigenvalues $L_{p}$.

The logarithm of analytic torsion $T(K)$ of $K$ can be defined as,
\begin{align}\label{eq:torsion}
\log T(K) = \frac{1}{2}{\sum\limits_{p = 0}^{n}{( - 1)^{p}p\zeta'_{p}(0)}}
\end{align}
where $\zeta'_{p}(0) = \zeta'_{p}(s=0)$ is the derivative of the Zeta function at $s=0$. Mathematically, if we let $|L_{p}|=\Pi_{\lambda_{i} > 0}\lambda_{i}$ be the product of all non-zero eigenvalues of $L_{p}$, we have $\zeta'_{p}(0)=-\log |L_{p}|$. Further, the logarithm of analytic torsion $T(K)$ can be rewritten as,
\begin{align}  \nonumber
\log T(K) = \frac{1}{2}\sum_{p = 0}^{n}(-1)^{p + 1}p\log |L_{p}|.
\end{align}
Note that the sum is from $0$ to $n$, the highest order of simplicial complex $K$.

For an $1$-dimension simplicial complex $K$, i.e., a graph, its analytic torsion $T(K)$ can be expressed as,
\begin{align}  \nonumber
T(K) = |L_{1}|^{\frac{1}{2}}.
\end{align}
Note that here $L_{1}=B_{1}^{T}B_{1}$ from Eq.(\ref{eq:L}). Further, for a $2$-dimension simplicial complex $K$, its analytic torsion $T(K)$ is,
\begin{align}  \nonumber
T(K) = \frac{|L_{1}|^{\frac{1}{2}}}{|L_{2}|}.
\end{align}
The analytic torsion is highly related to the dimension of the simplical complex. Figure \ref{fig:example of analytic torsion} illustrates the process to calculate the logarithm of analytic torsion for a $2$-dimension simplicial complex.

\subsection{Analytic torsion based graph neural networks}
In our TorGNN model, the analytic torsion is incorporated into graph neural network architecture by adding analytic torsion into message-passing process.

\begin{figure*}[h]
\begin{center}
\includegraphics[width=5.5in]{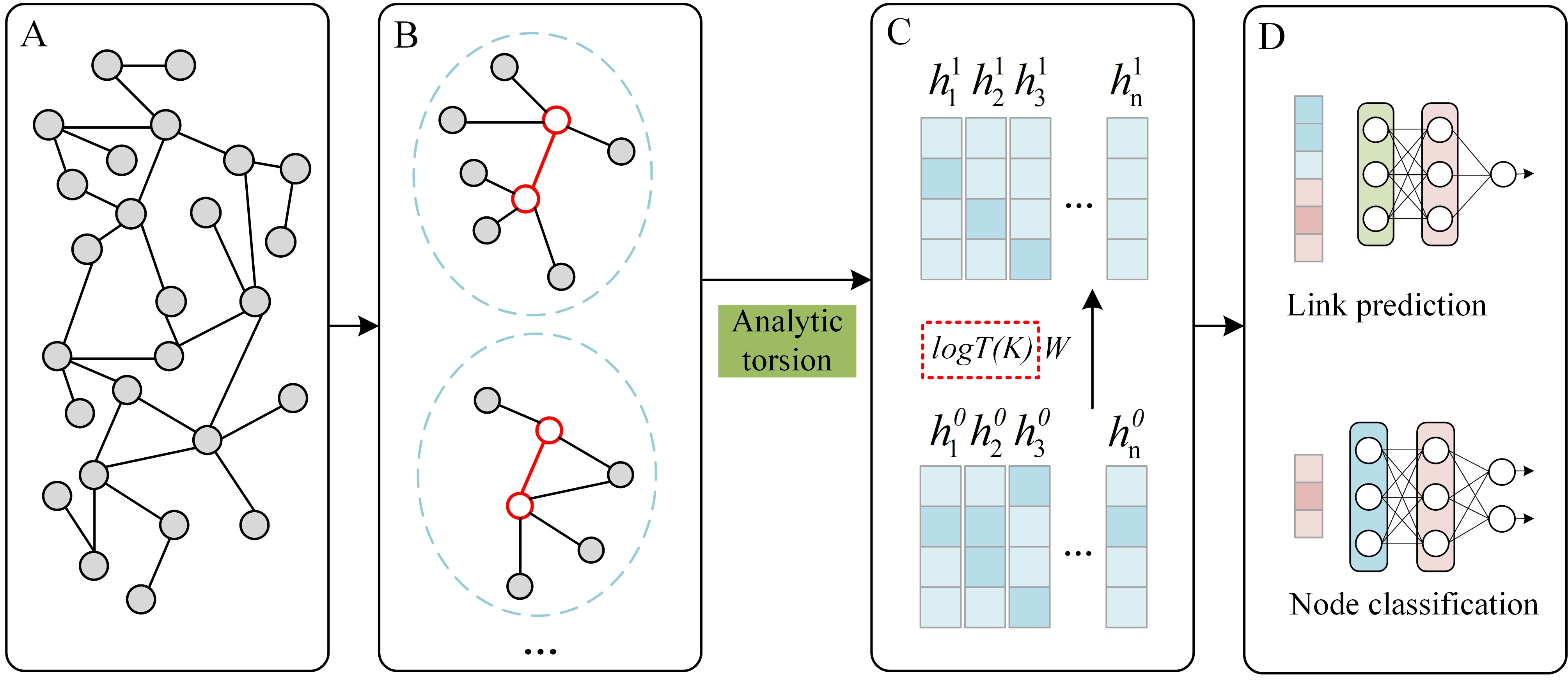}
\caption{\textbf{The overview of TorGNN architecture.} \textbf{A.} A large netwok. \textbf{B.} Local simplicial complex construction. \textbf{C.} Analytic torsion-based message-passing process. \textbf{D.} TorGNN architecture for link prediction task and node classification.}\label{fig:overview}
\end{center}
\end{figure*}

\textbf{Analytic torsion based message-passing process:}
An essential idea of our TorGNN is to aggregate node features by a analytic torsion-based edge weight as follows,
\begin{small}  
\begin{equation}\nonumber
h_{x}^{l} = \sigma\left( {\sum\limits_{y \in \mathcal{N}{(x)} \cup {\{ x\}}}{\frac{1}{\sqrt{d(x)} \sqrt{d(y)}} \left|\log T(K_{x,y})\right| W_{\rm GNN}h_{y}^{l - 1}}} \right),
\end{equation}
\end{small}

where $\mathcal{N}(x)$ is the neighbors of node $x$,  $d(x)$ and $d(y)$ represent node degree of nodes $x$ and $y$ respectively, $h_{x}^{l}$ and $h_{y}^{l - 1}$ are the node features of $x$ and $y$ after $l$ and $l-1$ (message-passing) iterations respectively, and $W_{\rm GNN}$ is the weight matrix to be learned.

More importantly, $K_{x,y}$ is the local simplicial complex constructed based on edge ${x, y}$. There are various ways to define this local simplicial complex. The most straightforward approach is to build a subgraph using the first-order neighbors. This subgraph contains vertices $x$, $y$, and all their neighbors together with all the edges among these vertices. If we allow each triangle in the subgraph to form a 2-simplex, a $2$-dimension $K_{x,y}$ can be obtained. Further, we can consider not only the direct neighbors of vertices $x$ and $y$, but also the indirect neighbors that within $l_{\rm sub}$ steps. An $p$-simplex $(p \leq n)$ is formed among $p+1$ vertices if each two of them form an edge. In general, our local simplicial complex $K_{x,y}$ depends on two parameters, i.e., step $l_{\rm sub}$ for neighbors and simplicial complex order $n$. The corresponding TorGNN model is denoted as ${\rm TorGNN}(l_{\rm sub}, n)$.

Our analytic torsion based message-passing process is illustrated in Figure \ref{fig:overview}. We consider ${\rm TorGNN}(1, 1)$ model, that is the local simplicial complex is constructed using $l_{\rm sub}=1$ and $n=1$. The logarithm of the analytic torsion for the local simplicial complex is used as a weight parameter in the message-passing process.

\textbf{TorGNN architectures:} Once we get the representation vector for each node, we can use the node representation for downstream tasks. In this section, we focus on link prediction tasks and node classification tasks.

The link prediction tasks are to calculate whether or not there is an edge between two nodes or the possibility of an edge. After the message-passing process, node representations $h_x$ and $h_y$ are obtained for nodes $x$ and $y$. The representation vector of the edge is then used as input for multilayer perceptron ($MLP$) to predict the possibility of the existence of the edge as follows,
\begin{align}  \nonumber
{\hat{p}}_{(x,y)} = {MLP}_{link}\left( \parallel \left( h_{x} + h_{y},h_{x}\odot h_{y},h_{x},~h_{y} \right) \right),
\end{align}
where $\parallel()$ is to concatenate the vectors within the bracket into a long vector, and $\odot$ is element-wise product. It is worth noting that various operations are used in the above equation to model the relationship between two nodes.

The node classification tasks are to predict the label of a node. In these tasks, we use the node representations learned from GNN part, and train them with a $MLP$ model with output the predicted label of the node. The process can be expressed as follows,
\begin{align}  \nonumber
{\hat{p}}_{x} = { MLP}_{node}\left( h_{x} \right)
\end{align}

Note that in both tasks, we use the cross-entropy as loss function. A detailed illustration of our TorGNN architectures are illustrated in Figure \ref{fig:overview}.
For more detailed parameter introduction of the model, please refer to the source code\footnote{A reference implementation of TorGNN may be found at  https://github.com/CS-BIO/TorGNN}.

\section{Experiments}
This section covers four tests, including a link prediction test on 16 networks, a node classification test on 3 networks, a representation learning test and a parameter analysis test. Our TorGNN model shows good performance on all tests.

\subsection{Datasets}
In this test, we utilize six types of datasets with a total of 19 networks to verify the performance of TorGNN. These six types of datasets include: 4 biomedical networks (HuRI-PPI \cite{luck2020reference}, ChG-Miner \cite{biosnapnets}, DisGeNET \cite{pinero2016disgenet}, Drugbank\_DTI \cite{wishart2018drugbank}), 4 social networks (lasftm\_asia \cite{rozemberczki2020characteristic}, twitch\_EN \cite{rozemberczki2019multiscale}, facebook\_comp \cite{rozemberczki2019gemsec}, facebook\_TV \cite{rozemberczki2019gemsec}), 2 collaboration networks (CA-HepTh \cite{leskovec2007graph}, CA-GrQc \cite{leskovec2007graph}), 3 internet peer-to-peer networks (p2p-Gnutella04 \cite{leskovec2007graph}, p2p-Gnutella05 \cite{leskovec2007graph}, p2p-Gnutella06 \cite{leskovec2007graph}), 3 autonomous systems networks (as20000102 \cite{leskovec2005graphs}, oregon1\_010331 \cite{leskovec2005graphs}, oregon1\_010407 \cite{leskovec2005graphs}), and 3 citation networks (Citeseer \cite{yang2016revisiting}, Cora \cite{yang2016revisiting}, Pubmed \cite{yang2016revisiting}). Note that biomedical networks, social networks, collaboration networks, internet peer-to-peer networks and autonomous systems networks are mainly used for link prediction tasks. See Table \ref{tab:Statistics of networks on link prediction} for details of these dataset. The citation networks are mainly used for node classification tasks. See Table \ref{tab:Statistics of networks on node classification} for details of the dataset.
\begin{table}[ht]
\centering
\caption{Details of 16 network data on link prediction tasks.}\label{tab:Statistics of networks on link prediction}
\scalebox{0.95}{
\begin{tabular}{llllll}
\hline
Categories& Networks & Nodes & Edges & Density \\
 \hline
\multirow{4}{*}{\makecell[l]{Biomedical \\networks}} &HuRI-PPI	&5604	&23322	 &0.15\%\\
&ChG-Miner	&7341	&15138	&0.06\%\\
&DisGeNET	&19783	&81746	&0.04\%\\
&Drugbank\_{DTI}	&12566	&18866	&0.02\%\\
\hline
\multirow{4}{*}{Social networks} &lasftm\_asia	&7624	&27806	&0.10\%\\
&twitch\_EN &7126	&35324 &0.14\%\\
&facebook\_comp &14113	&52310 &0.05\%\\
&facebook\_TV &3892	&17262 &0.23\%\\
\hline
\multirow{2}{*}{\makecell[l]{Collaboration \\networks}} &CA-HepTh	&9877	&25998		 &0.05\%\\
&CA-GrQc	&5242	&14496	&0.11\%\\
\hline
\multirow{3}{*}{\makecell[l]{Internet \\peer-to-peer \\networks}} &p2p-Gnutella04	&10876	 &39994	&0.07\%\\
&p2p-Gnutella05	&8846	&31839	&0.08\%\\
&p2p-Gnutella06	&8717	&31525	&0.08\%\\
\hline
\multirow{3}{*}{\makecell[l]{Autonomous \\systems \\networks}} &as20000102	&6474	&13895	 &0.07\%\\
&oregon1\_010331	&10670	&22002	&0.04\%\\
&oregon1\_010407	&10729	&21999	&0.04\%\\
\hline
\end{tabular}}
\end{table}

\begin{table}[ht]
\centering
\caption{Details of 3 network data on node classification tasks.}\label{tab:Statistics of networks on node classification}
\begin{tabular}{lllllll}
\hline
Networks & Nodes & Edges & Density  & Classes  & Feature \\
 \hline
Citeseer 	&3327	&4732	&0.09\%	&6	 &3703\\
Cora	&2708	&5429	&0.15\%	&7	&1433\\
Pubmed	&19717	&44338	&0.02\%	&3	&500\\

\hline
\end{tabular}
\end{table}

\subsection{Baselines}
In link prediction tasks, we compare TorGNN against state-of-the-art GNN methods and network embedding methods. We adopt a total of 5 GNN models, including LightGCN \cite{he2020lightgcn}, SkipGNN \cite{huang2020skipgnn}, KGIN \cite{wang2021learning}, GCN \cite{welling2016semi} and GAT \cite{velivckovic2017graph}. The network embedding method is a kind of representation learning methods, and it's main goal is to reduce high-dimensional representation vectors of nodes, edges, or subgraphs into low-dimensional vectors. We select three classic network embedding methods, including DeepWalk \cite{perozzi2014deepwalk}, LINE \cite{tang2015line} and SDNE \cite{wang2016structural}.

In node classification tasks, we select 9 models as the comparison methods, namely ManiReg \cite{belkin2006manifold}, SemiEmb \cite{weston2008deep}, DeepWalk \cite{perozzi2014deepwalk}, Planetoid \cite{yang2016revisiting}, GCN \cite{welling2016semi}, DCNN \cite{atwood2016diffusion}, SAGE \cite{hamilton2017inductive}, N-GCN \cite{abu2020n} and N-SAGE \cite{abu2020n}. These 9 methods cover various model types that can be applied to node classification tasks, including graph neural network, network embedding model, traditional machine model, etc. In this way,  the performance of our TorGNN model can be evaluated more comprehensively.

\begin{table*}[ht]
\centering
\caption{The comparison of different models on link prediction based on 16 datasets. The measurements are AUC and AUPR. }\label{tab:Link prediction}
\begin{tabular}{lllllllllll}
\hline
&Networks	&TorGNN	&LightGCN	&SkipGNN	&KGIN	&GCN	&GAT	&DeepWalk	&LINE	 &SDNE\\
\hline
AUC &HuRI-PPI	&\textbf{0.9369}	&0.8170	&0.9119	&0.9007	&0.9164	&0.8994	&0.7294	&0.8223	 &0.9243\\
&ChG-Miner	&\textbf{0.9583}	&0.7457	&0.9526	&0.9493	&0.9352	&0.9514	&0.8093	&0.7036	 &0.6108\\
&DisGeNET	&\textbf{0.9868}	&0.8597	&0.9145	&0.9154	&0.9723	&0.9829	&0.6922	&0.8492	 &0.9586\\
&Drugbank\_DTI	&\textbf{0.9651}	&0.6821	&0.8946	&0.9453	&0.9234	&0.9476	&0.8572	&0.6312	 &0.8522\\
&lasftm\_asia	&\textbf{0.9204}	&0.8787	&0.8117	&0.9125	&0.8693	&0.8619	&0.7312	&0.8049	 &0.8952\\
&twitch\_EN	&\textbf{0.9159}	&0.8071	&0.8640	&0.8807	&0.8965	&0.8863	&0.6823	&0.7836	 &0.8627\\
&facebook\_comp	&\textbf{0.8974}	&0.8793	&0.7823	&0.9146	&0.8614	&0.8598	&0.7525	&0.6756	 &0.7733\\
&facebook\_TV	&\textbf{0.9500}	&0.9164	&0.7941	&0.8951	&0.9146	&0.9165	&0.8637	&0.6841	 &0.8145\\
&CA-HepTh	&\textbf{0.9476}	&0.6004	&0.7859	&0.8117	&0.9206	&0.9185	&0.7454	&0.6285	 &0.8514\\
&CA-GrQc	&\textbf{0.9599}	&0.8304	&0.8062	&0.8084	&0.9430	&0.9420	&0.8224	&0.7307	 &0.8785\\
&p2p-Gnutella04	&\textbf{0.9060}	&0.6232	&0.8034	&0.8898	&0.8901	&0.8816	&0.6175	&0.5980	 &0.7883\\
&p2p-Gnutella05	&\textbf{0.8956}	&0.6037	&0.8074	&0.8871	&0.8847	&0.8785	&0.6353	&0.5945	 &0.8051\\
&p2p-Gnutella06	&\textbf{0.9051}	&0.6240	&0.8112	&0.8864	&0.8872	&0.8828	&0.6490	&0.5687	 &0.8003\\
&as20000102	&0.9292	&0.7834	&0.9125	&\textbf{0.9334}	&0.9120	&0.8926	&0.8276	&0.8340	 &0.8902\\
&oregon1\_010331	&\textbf{0.9602}	&0.7913	&0.9544	&0.9450	&0.9516	&0.9338	&0.8165	 &0.8812	 &0.9219\\
&oregon1\_010407	&\textbf{0.9603}	&0.8060	&0.9492	&0.9416	&0.9518	&0.9594	&0.8142	 &0.8657	 &0.9172\\
\cline{2-11}

&Average	&\textbf{0.9372} &0.7655	&0.8597	&0.9011	&0.9144	&0.9107	&0.7529	&0.7285	 &0.8465\\
\hline
AUPR &HuRI-PPI	&\textbf{0.9424}	&0.8589	&0.9182	&0.9006	&0.9189	&0.8965	&0.6958	&0.8520	 &0.9324\\
&ChG-Miner	&\textbf{0.9606}	&0.8006	&0.9524	&0.9493	&0.9409	&0.9499	&0.8267	&0.7514	 &0.6114\\
&DisGeNET	&\textbf{0.9882}	&0.8915	&0.9271	&0.9154	&0.9785	&0.9849	&0.6846	&0.8477	 &0.9554\\
&Drugbank\_DTI	&\textbf{0.9664}	&0.7593	&0.6764	&0.9452	&0.9371	&0.9533	&0.8729	&0.7044	 &0.8672\\
&lasftm\_asia	&\textbf{0.9311}	&0.9028	&0.8269	&0.9124	&0.8810	&0.8810	&0.7365	&0.8395	 &0.9127\\
&twitch\_EN	&\textbf{0.9265}	&0.8573	&0.8750	&0.8807	&0.9051	&0.8899	&0.6729	&0.8134	 &0.8642\\
&facebook\_comp	&0.9112	&0.9045	&0.7993	&\textbf{0.9146}	&0.8761	&0.871	&0.7517	&0.6956	 &0.8092\\
&facebook\_TV	&\textbf{0.9569}	&0.9359	&0.8116	&0.8951	&0.9256	&0.9265	&0.8672	&0.7461	 &0.8517\\
&CA-HepTh	&\textbf{0.9487}	&0.5991	&0.7889	&0.7794	&0.9169	&0.9077	&0.7313	&0.6778	 &0.8614\\
&CA-GrQc	&\textbf{0.9626}	&0.8275	&0.8228	&0.7769	&0.9427	&0.9419	&0.8469	&0.7863	 &0.8845\\
&p2p-Gnutella04	&\textbf{0.9027}	&0.6884	&0.7660	&0.8898	&0.8901	&0.8843	&0.6166	&0.5967	 &0.7735\\
&p2p-Gnutella05	&\textbf{0.8892}	&0.6764	&0.7748	&0.8870	&0.8804	&0.8751	&0.6514	&0.6177	 &0.7838\\
&p2p-Gnutella06	&\textbf{0.8996}	&0.6918	&0.7763	&0.8862	&0.8830	&0.8814	&0.6457	&0.5976	 &0.7718\\
&as20000102	&\textbf{0.9383}	&0.8403	&0.9333	&0.9334	&0.9290	&0.9090	&0.8157	&0.8742	 &0.9159\\
&oregon1\_010331	&\textbf{0.9638}	&0.8397	&0.9601	&0.9450	&0.9596	&0.9448	&0.8088	 &0.9042	 &0.9354\\
&oregon1\_010407	&\textbf{0.9649}	&0.8573	&0.9604	&0.9415	&0.9596	&0.9448	&0.7888	 &0.8987	 &0.9335\\
\cline{2-11}
&Average	&\textbf{0.9319}	&0.9008	&0.8355	&0.8685	&0.8649	&0.8477	&0.7731	&0.7575	 &0.7992\\
\hline
\end{tabular}
\end{table*}

\subsection{Performance on link prediction tasks}
Five types of datasets, including a total of 16 network data, are used for the evaluation of the performance of our TorGNN. For each network, we randomly sample the same number of negative and positive samples, and then divide them into training set, validation set and test set with a ratio of 7:1:2. The area under the receiver operating characteristic curve (AUC) and the area under the precision-recall curve (AUPR) are used for evaluating the performance of each model. This process is repeated 10 times, and the average value is taken as the final result. We use batch size 128 with Adam optimizer and run TorGNN model in PyTorch. For the learning rate, after adjustment, it is found that the learning rate of 5e-4 is the most suitable for all 16 networks.

Table \ref{tab:Link prediction} presents the prediction results on the link prediction tasks. Overall, our TorGNN model outperforms other models on most datasets, and the average AUC and AUPR on 16 networks are 2.49\% and 3.45\% higher than the second-ranked model, respectively. It is worth noting that the performance of the TorGNN model on all data sets is better than that of the 3 network embedding models (DeepWalk, LINE and SDNE), which shows that the TorGNN model is not only a good graph neural network model, but also an excellent network embedding models. Although the AUC value of the TorGNN model on the ``as20000102" network and the AUPR value on the ``facebook\_comp" network did not reach the highest value, the AUPR value of the TorGNN model on the ``as20000102" network and the AUC value on the ``facebook\_comp" network are still the best. Note that our tasks are from various types of networks, including biomedical networks, collaboration networks and  internet peer-to-peer networks, the great performance of our TorGNN shows that it has a strong generalization capability for network/graph data.

\subsection{Performance on node classification tasks.}
Three networks, including Citeseer, Cora and Pubmed, are used in the evaluation of the performance of TorGNN model. Following the general way to set up training, verification and test sets of these three datasets \cite{rong2019dropedge}, we select 500 nodes in each data set as the verification set, 1000 nodes as the test set, and the remaining nodes as the training set. The accuracy is used to evaluate the performance of TorGNN. Each step is run 10 times, and the average is taken as the final result. We use Adam optimizer of learning rate 0.02 and run TorGNN model in PyTorch.

\begin{figure}[ht]
\begin{center}
\includegraphics[width=2.5in]{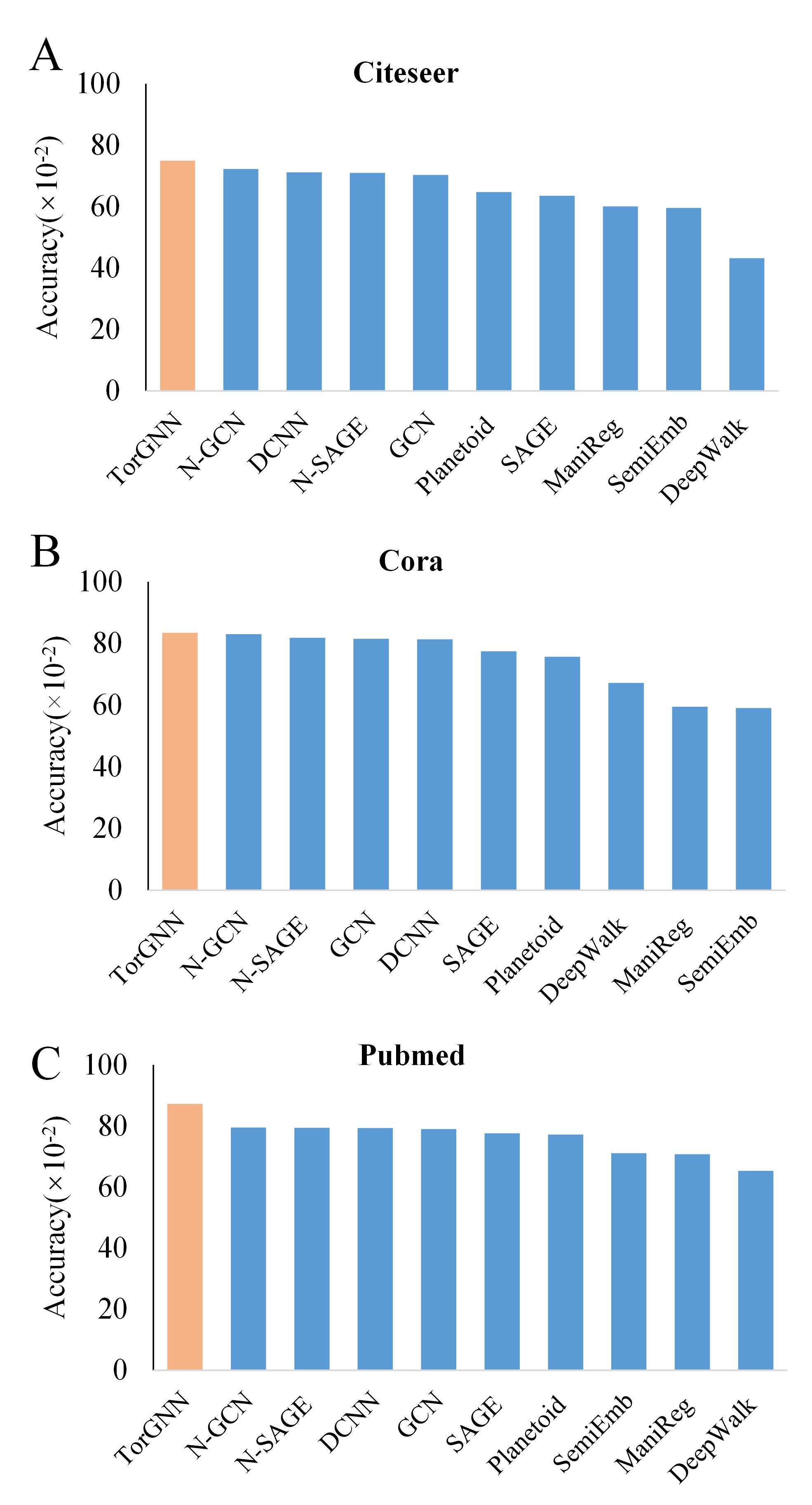}
\caption{\textbf{The comparison of different models on node classification performance based on three citation datasets. )}. \textbf{A}, \textbf{B}, and \textbf{C} illustrate the comparison of accuracy for various models on Citeseer, Cora and Pubmed datasets, respectively. }\label{fig:classification}
\end{center}
\end{figure}

Figure \ref{fig:classification} shows the results of the TorGNN model and the comparison methods for node classification tasks on three datasets. Our TorGNN model has achieved the best results, and it is significantly better than the second-ranked model N-GCN. It is worth noting that N-GCN and N-SAGE are improved models based on GCN and SAGE. Specifically, they use a data perturbation method to prevent overfitting of the traditional model. As can be seen in Figure \ref{fig:classification}, the performance of N-GCN and N-SAGE models is superior to the traditional GCN and SAGE models. Nevertheless, these models are still slightly inferior to our TorGNN model, which shows that the advantages of our TorGNN model over the traditional GNNs are not only reflected in the overall performance, but also in preventing overfitting. The reason may be that the analytic torsion of the local structure plays a restrictive role in the adjustment of the parameters in GNNs and reduces overfitting.

\subsection{Performance on representation learning}
\begin{figure}[ht]
\begin{center}
\includegraphics[width=3.2in]{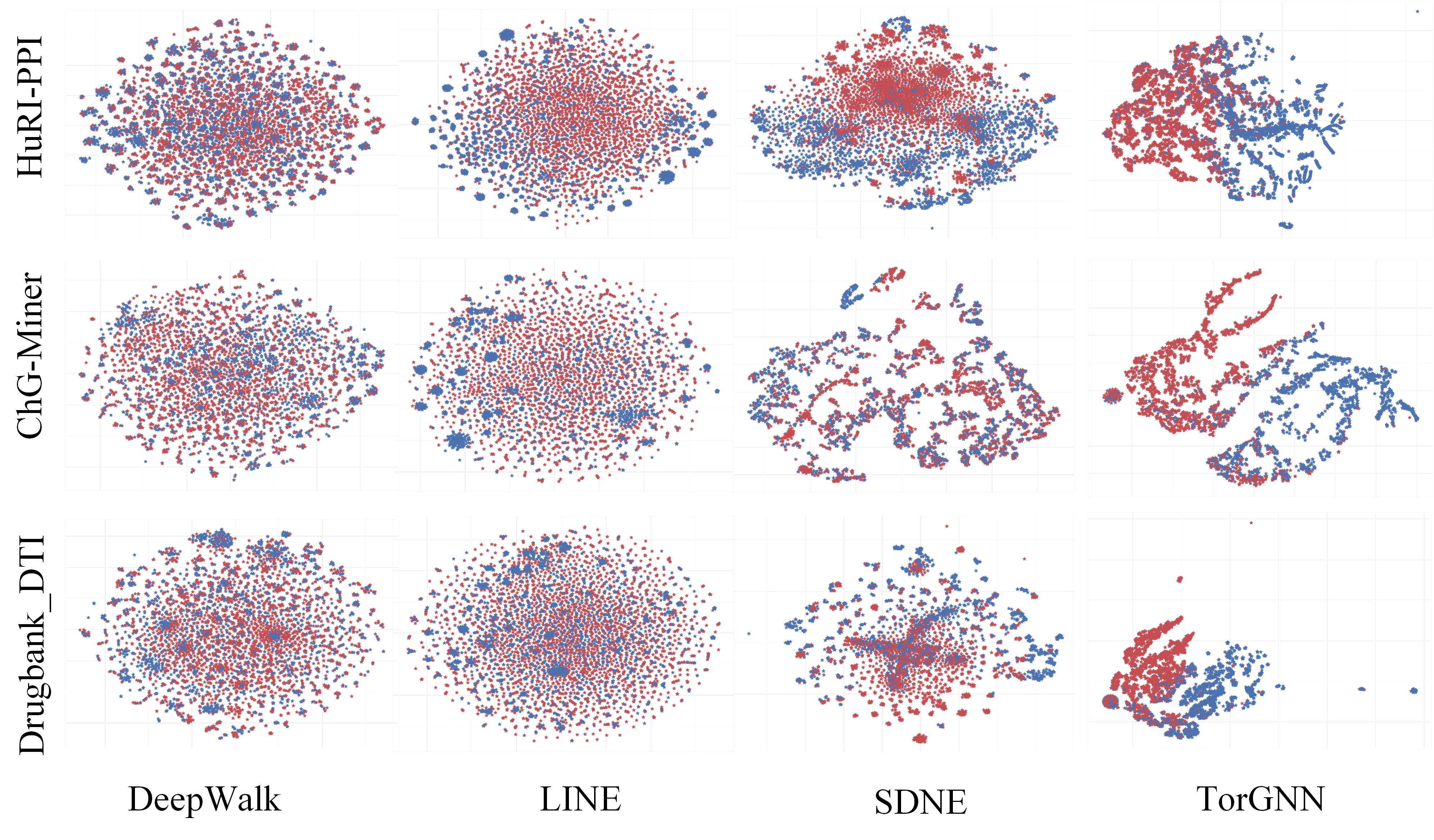}
\caption{\textbf{The comparison of the representation learning capability of different models on HuRI-PPI, ChG-Miner and Drugbank\_DTI network datasets.} The representation vectors of each nodes in the test datasets are projected into 2D spaces by t-SNE. The red and blue points represent node pairs without and with link relationships, respectively. Four network embedding methods are considered in our comparison.} \label{fig:representation learning}
\end{center}
\end{figure}

To further illustrate the representation learning ability of our TorGNN, we set up a visualization test. First, we get the representation vectors for node pairs in the test sets. Then, we use t-SNE \cite{van2008visualizing} to project the high-dimensional representation vectors to 2D space, where the representation vectors are mainly divided into two categories, positive samples (node pairs with link relationship) and negative samples (node pairs without link relationship). Finally, these representation vectors in 2D space are illustrated in Figure \ref{fig:representation learning} (red and blue points represent negative and positive samples, respectively). Three datasets of HuRI-PPI, ChG-Miner and Drugbank\_DTI are used in our test. At the same time, we choose DeepWalk, LINE and SDNE three network embedding methods for comparison. It can be seen from Figure \ref{fig:representation learning} that our TorGNN model has obvious advantages in distinguishing node pairs with and without link relationship, which shows that our TorGNN model has a better ability in representation learning than traditional network embedding model.

\subsection{Influence of local simplicial complex}
In order to analyze the impact of the number of neighbor layers and the highest order of the local simplicial complex on ${\rm TorGNN}(l_{\rm sub}, n)$ , we set up the following parameter analysis model:
\begin{itemize}
\item TorGNN(1,1): $l_{sub}$=1, $n$=1. It means that TorGNN only considers first-order neighbors when constructed local simplicial complex, the highest order of the simplicial complex is only up to 1, that is, only edges are considered.
\item TorGNN(1,2): $l_{sub}$=1, $n$=2. It means that TorGNN only considers first-order neighbors when constructed local simplicial complex, the highest order of the simplicial complex is up to 2, that is, triangles (2-simplices) are considered.
\item TorGNN(2,1): $l_{sub}$=2, $n$=1. It means that TorGNN considers second-order neighbors when constructed local simplicial complex, the highest order of the simplicial complex is only up to 1, that is, only edges are considered.
\end{itemize}

Table \ref{tab:Effect pf parameter on link prediction tasks} and Table \ref{tab:Effect pf parameter on node classification tasks} show the results of these three models on link prediction tasks and node classification tasks. It can be seen that the performance of the three models are relatively stable. The differences between the three models for both AUC and AUPR are less than 0.02 for all the test cases, except ``drugbank\_DTI", ``facebook\_comp" and ``p2p-Gnutella06". Comparably speaking, TorGNN(2,1) model has the best overall performance. This may due to the reason that the corresponding local simplicial complex of TorGNN(2,1) has a better representability of the local structures. Note that with $l_{sub}=2$, the local simplicial complex contains all second-order neighbors and is a much larger structure. Computationally, more efficient algorithms will be needed for the fast evaluation of the analytical torsion if large-sized simplicial complex is used.


\begin{table}[ht]
\caption{Effect of parameters on link prediction tasks of ${\rm TorGNN}(l_{\rm sub}, n)$.}\label{tab:Effect pf parameter on link prediction tasks}
\begin{threeparttable}
\resizebox{\columnwidth}{!}{
\begin{tabular}{llllllll}
\hline
\multirow{2}{*}{Networks}&\multicolumn{2}{l}{TorGNN(1,1)} &\multicolumn{2}{l}{TorGNN(1,2)} &\multicolumn{2}{l}{TorGNN(2,1)}\\
& AUC & AUPR & AUC & AUPR & AUC & AUPR \\
 \hline
HuRI-PPI	&\textbf{0.9369}	&\textbf{0.9424}	&0.9308	&0.9377	&0.9357	&0.9370\\
ChG-Miner	&0.9533	&0.9570	&0.9475	&0.9524	&\textbf{0.9583}	&\textbf{0.9606}\\
DisGeNET	&\textbf{0.9868}	&\textbf{0.9882}	&0.9840	&0.9857 &NA\tnote{1}	 &NA\tnote{1}	 \\
drugbank\_DTI	&0.9380	&0.9469	&0.9306	&0.9411	&\textbf{0.9651}	&\textbf{0.9664}\\
lasftm\_asia	&0.9152	&0.9261	&0.8997	&0.9150	&\textbf{0.9204}	&\textbf{0.9311}\\
twitch\_EN	&\textbf{0.9159}	&\textbf{0.9265}	&0.9066	&0.9168	&NA\tnote{1}&NA\tnote{1}	 \\
facebook\_comp	&\textbf{0.8974}	&\textbf{0.9112}	&0.8865	&0.9012	&0.8573	&0.8682\\
facebook\_TV	&0.9488	&0.9564	&NA\tnote{1}	&NA\tnote{1}	&\textbf{0.9500}	 &\textbf{0.9569}\\
CA-HepTh	&0.9427	&0.9430	&0.9289	&0.9288	&\textbf{0.9476}	&\textbf{0.9487}\\
CA-GrQc	&0.9573	&0.9597	&0.9328	&0.9373	&\textbf{0.9599}	&\textbf{0.9626}\\
p2p-Gnutella04	&0.9024	&0.9026	&0.8898	&0.8923	&\textbf{0.9060}	&\textbf{0.9027}\\
p2p-Gnutella05	&0.8883	&0.8858	&0.8843	&0.8836	&\textbf{0.8956}	&\textbf{0.8892}\\
p2p-Gnutella06	&0.8464	&0.8978	&0.8913	&0.8908	&\textbf{0.9051}	&\textbf{0.8996}\\
as20000102	&0.9274	&0.9364	&0.9216	&0.9335	&\textbf{0.9292}	&\textbf{0.9383}\\
oregon1\_010331	&0.9590	&0.9642	&0.9505	&0.9593	&\textbf{0.9602}	&\textbf{0.9638}\\
oregon1\_010407	&\textbf{0.9603}	&\textbf{0.9649}	&0.9524	&0.9600	&0.9590	&0.9637\\
\hline
\end{tabular}}
\begin{tablenotes}
\footnotesize
\item[1] NA indicates that the model requires too much memory or time.
\end{tablenotes}
\end{threeparttable}
\end{table}

\begin{table}[ht]
\centering
\caption{Effect of parameters on node classification tasks of ${\rm TorGNN}(l_{\rm sub}, n)$.}\label{tab:Effect pf parameter on node classification tasks}
\begin{tabular}{llllllll}
\hline
\multirow{2}{*}{Networks}&\multicolumn{3}{l}{Accuracy}\\
\cline{2-4}
& TorGNN(1,1) & TorGNN(1,2) & TorGNN(2,1)\\
 \hline
Citeseer	&0.748	&0.746	&\textbf{0.749}\\
Cora	&0.832	&0.823	&\textbf{0.834}\\
Pubmed	&0.865	&\textbf{0.872}	&0.868\\
\hline
\end{tabular}
\end{table}

\section{Conclusion}
The incorporation of geometric and topological information into graph neural architecture remains to be a key issue for geometric deep learning models, in particular graph neural network models.

In this paper, we propose an analytic torsion enhanced graph neural network (TorGNN) model, which mainly uses analytic torsion to capture the complexity of the structures, and then incorporate it into message-passing process to improve the performance of the traditional graph neural network models. From the tasks of link prediction and node classification, our TorGNN is found to be not only better than traditional GNNs models, but also better than other graph deep learning models.

\section*{Acknowledgements}
This work was supported in part by National Natural Science Foundation of China (NSFC grant no. 61873089, 62032007), Nanyang Technological University Startup Grant (grant no. M4081842), Singapore Ministry of Education Academic Research fund (grant no. Tier 1 RG109/19, MOE-T2EP20120-0013, MOE-T2EP20220-0010) and China Scholarship Council (CSC grant no.202006130147).

\bibliographystyle{IEEEtran}
\bibliography{example_paper}

\end{document}